\documentclass[journal]{IEEEtran}
\usepackage{enumitem}
\usepackage{subcaption}
\usepackage{booktabs}
\usepackage{diagbox}
\usepackage[pdftex]{graphicx}
\PassOptionsToPackage{hyphens}{url}\usepackage{hyperref}

\begin{document}
%
\title{EuroSAT: A Novel Dataset and Deep Learning Benchmark for Land Use and Land Cover Classification}
%
%
%

\author{
   \fontsize{12pt}{0}\fontfamily{phv}\selectfont 
        ~ Patrick Helber\textsuperscript{1,2}
        ~ Benjamin Bischke\textsuperscript{1,2}
        ~ Andreas Dengel\textsuperscript{1,2} 
        ~ Damian Borth\textsuperscript{2} ~  \\ ~ \\
    \fontsize{10pt}{0}\fontfamily{phv}\selectfont 
    \hspace{-0.25em}
    $^1$TU Kaiserslautern, Germany ~~~~ $^2$German Research Center for Artificial Intelligence (DFKI), Germany  \\ 
    \fontsize{10pt}{0}\fontfamily{phv}\selectfont 
     \{Patrick.Helber, Benjamin.Bischke, Andreas.Dengel, Damian.Borth\}@dfki.de  \\ 
}

\maketitle

\begin{abstract}
In this paper, we address the challenge of land use and land cover classification using Sentinel-2 satellite images. The Sentinel-2 satellite images are openly and freely accessible provided in the Earth observation program Copernicus. We present a novel dataset based on Sentinel-2 satellite images covering 13 spectral bands and consisting out of 10 classes with in total 27,000 labeled and geo-referenced images. We provide benchmarks for this novel dataset with its spectral bands using state-of-the-art deep Convolutional Neural Network (CNNs). With the proposed novel dataset, we achieved an overall classification accuracy of 98.57\%. The resulting classification system opens a gate towards a number of Earth observation applications. We demonstrate how this classification system can be used for detecting land use and land cover changes and how it can assist in improving geographical maps. The geo-referenced dataset EuroSAT is made publicly available at https://github.com/phelber/eurosat.
\end{abstract}


%
\IEEEpeerreviewmaketitle

\section{Introduction}
%
%
%
%
\IEEEPARstart{W}e are currently at the edge of having public and continuous access to satellite image data for Earth observation. Governmental programs such as ESA's Copernicus and NASA's Landsat are taking significant efforts to make such data freely available for commercial and non-commercial purpose with the intention to fuel innovation and entrepreneurship. With access to such data, applications in the domains of agriculture, disaster recovery, climate change, urban development, or environmental monitoring
can be realized~\cite{ponti2016precision, bischke2017multimediaSatelliteSolution, bischke2016contextual, bischke2017multimediaSatellite}. However, to fully utilize the data for the previously mentioned
domains, first satellite images must be processed and transformed into structured 
semantics~\cite{huang2018opensarship}. One type of such fundamental semantics is \textit{Land Use and Land Cover Classification}~\cite{basu2015deepsat, yang2010bag}. The aim of land use and land cover classification is to automatically provide labels describing the represented physical land type or how a land area is used (e.g., residential, industrial).

\begin{figure}[t]
	\centering
	\includegraphics[scale=0.6]{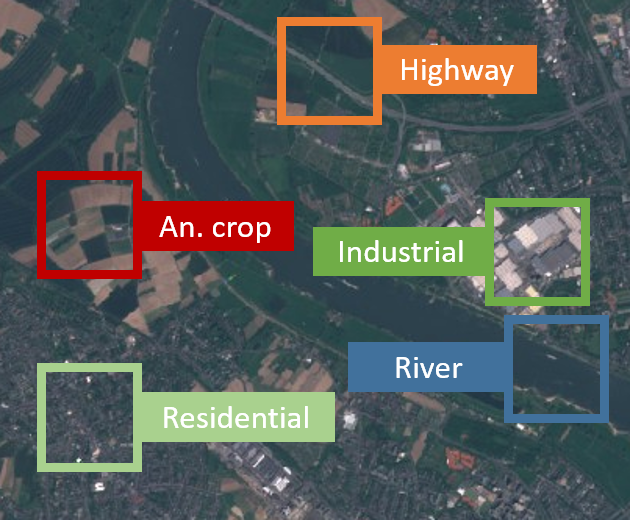}
	\caption{Land use and land cover classification based on Sentinel-2 satellite images. Patches are extracted with the purpose to identify the shown class. This visualization highlights the classes annual crop, river, highway, industrial buildings and residential buildings.}
	\label{fig:overview}
\end{figure}

As often in supervised machine learning, the performance of classification systems strongly depends on the availability of high-quality datasets with a suitable set of classes~\cite{ILSVRC15}. 
In particular when considering the recent success of deep \textit{Convolutional Neural 
Networks (CNN)}~\cite{krizhevsky2012imagenet}, it is crucial to have large quantities of training data available to train such a network. Unfortunately, current land use and land cover datasets are small-scale or rely on data sources which do not allow the mentioned domain applications. 

In this paper, we propose a novel satellite image dataset for the task of land use and land cover classification. The proposed EuroSAT dataset consists of 27,000 labeled images with 10 different land use and land cover classes. A significant difference to previous datasets is that the presented satellite image dataset is multi-spectral covering 13 spectral bands in the visible, near infrared and short wave infrared part of the spectrum. In addition, the proposed dataset is georeferenced and based on openly and freely accessible Earth observation data allowing a unique range of applications. The labeled dataset EuroSAT is made publicly available at https://github.com/phelber/eurosat. 
\begin{figure*}
	\centering
	\includegraphics[width=0.9\linewidth]{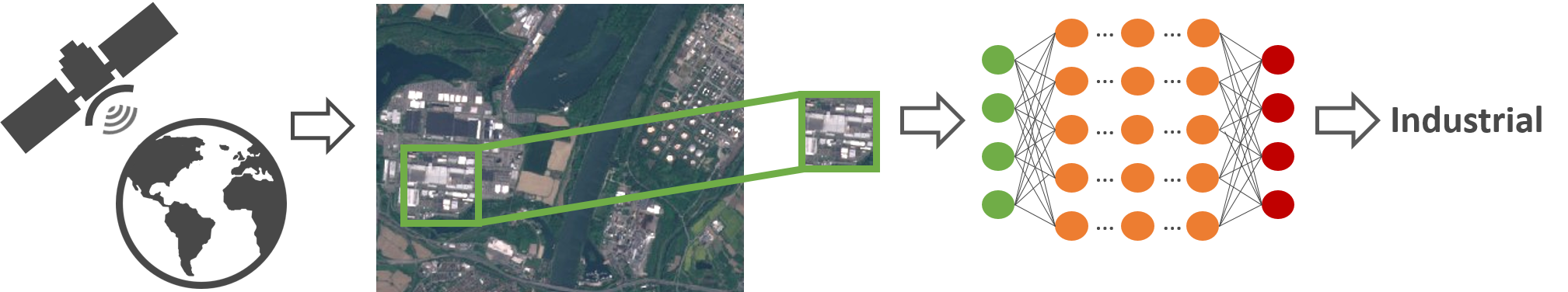}
	\caption{This illustration shows an overview of the patch-based land use and land cover classification process using satellite images. A satellite scans the Earth to acquire images of it. Patches extracted out of these images are used for classification. The aim is to automatically provide labels describing the represented physical land type or how the land is used. For this purpose, an image patch is feed into a classifier, in this illustration a neural network, and the classifier outputs the class shown on the image patch.}
	\label{fig:overview}
\end{figure*}
Further, we provide a full benchmark demonstrating a robust classification performance which is the basis for developing applications for the previously mentioned domains. We outline how the classification model can be used for detecting land use or land cover changes and how it can assist in improving geographical maps. 

We provide this work in the context of the recently published EuroSAT dataset, which can be used similar to~\cite{ni2015large} as a basis for a large-scale training of deep neural networks for the task of satellite image classification.

In this paper, we make the following \textbf{contributions}:
\begin{quote}
\begin{itemize}
\item We introduce the first large-scale patch-based land use and land cover classification dataset based on Sentinel-2 satellite images. Every image in the dataset is labeled and geo-referenced. We release the RGB and the multi-spectral (MS) version of the dataset.
\item We provide benchmarks for the proposed EuroSAT dataset using Convolutional Neural Networks (CNNs).
\item We evaluate the performance of each spectral band of the Sentinel-2 satellite for the task of patch-based land use and land cover classification.
\end{itemize}
\end{quote}

\section{Related Work}

In this section, we review previous studies in land use and land cover classification. In this context, we present remotely sensed aerial and satellite image datasets. Furthermore, we present state-of-the-art image classification methods for land use and land cover classification.


\subsection{Classification Datasets}

The classification of remotely sensed images is a challenging task. The progress of classification in the remote sensing area has particularly been inhibited due to the lack of reliably labeled ground truth datasets. A popular and intensively studied~\cite{castelluccio2015land, nogueira2017towards, penatti2015deep, xia2016aid, yang2010bag} remotely sensed image classification dataset known as \textit{UC Merced} (UCM) land use dataset was introduced by Yang et al.~\cite{yang2010bag}. The dataset consists of 21 land use and land cover classes. Each class has 100 images and the contained images measure 256x256 pixels with a spatial resolution of about 30 cm per pixel. All images are in the RGB color space and were extracted from the USGS National Map Urban Area Imagery collection, i.e. the underlying images were acquired from an aircraft. Unfortunately, a dataset with 100 images per class is small-scale. Trying to enhance the dataset situation, various works used commercial Google Earth images to manually create novel datasets~\cite{sheng2012high, xia2016aid, xia2010structural, zhao2016feature} such as the two benchmark datasets PatternNet~\cite{zhou2018patternnet} and NWPU-RESISC45~\cite{cheng2017remote}. The datasets are based on very-high-resolution images with a spatial resolution of up to 30 cm per pixel. Since the creation of a labeled dataset is extremely time-consuming, these datasets consist likewise of only a few hundred images per class. One of the largest datasets is the \textit{Aerial Image Dataset (AID)}. AID consists of 30 classes with 200 to 400 images per class. The 600x600 high-resolution images were also extracted from Google Earth imagery.

Compared to the EuroSAT dataset presented in this work, the previously listed datasets rely on commercial very-high-resolution and preprocessed images. The fact of using commercial and preprocessed very-high-resolution image data makes these datasets unsatisfying for real-world Sentinel-2 Earth observation applications as proposed in this work. Furthermore, while these datasets put a strong focus on strengthening the number of covered classes, the datasets suffer from a low number of images per class. The fact of a spatial resolution of up to 30 cm per pixel, with the possibility to identify and distinguish classes like churches, schools etc., make the presented datasets difficult to compare with the dataset proposed in this work.


A study closer to our work, provided by Penatti et al. ~\cite{penatti2015deep}, analyzed remotely sensed satellite images with a spatial resolution of 10 meters per pixel to classify coffee crops. Based on these images, Penatti et al.~\cite{penatti2015deep} introduced the \textit{Brazillian Coffee Scene (BCS) dataset}. The dataset covers the two classes coffee crop and non-coffee crop. Each class consists of 1,423 images. The images consist of a red, green and near-infrared band. 

Similar to the proposed EuroSAT dataset, Basu et al.~\cite{basu2015deepsat} introduced the \textit{SAT-6} dataset relying on aerial images. This dataset has been extracted from images with a spatial resolution of 1 meter per pixel. The image patches are created using images from the National Agriculture Imagery Program (NAIP). SAT-6 covers the 6 different classes: barren land, trees, grassland, roads, buildings and water bodies. The proposed patches have a size of 28x28 pixels per image and consist of a red, green, blue and a near-infrared band.

\begin{figure}[t!]
	\centering
	\includegraphics[width=.8\linewidth]{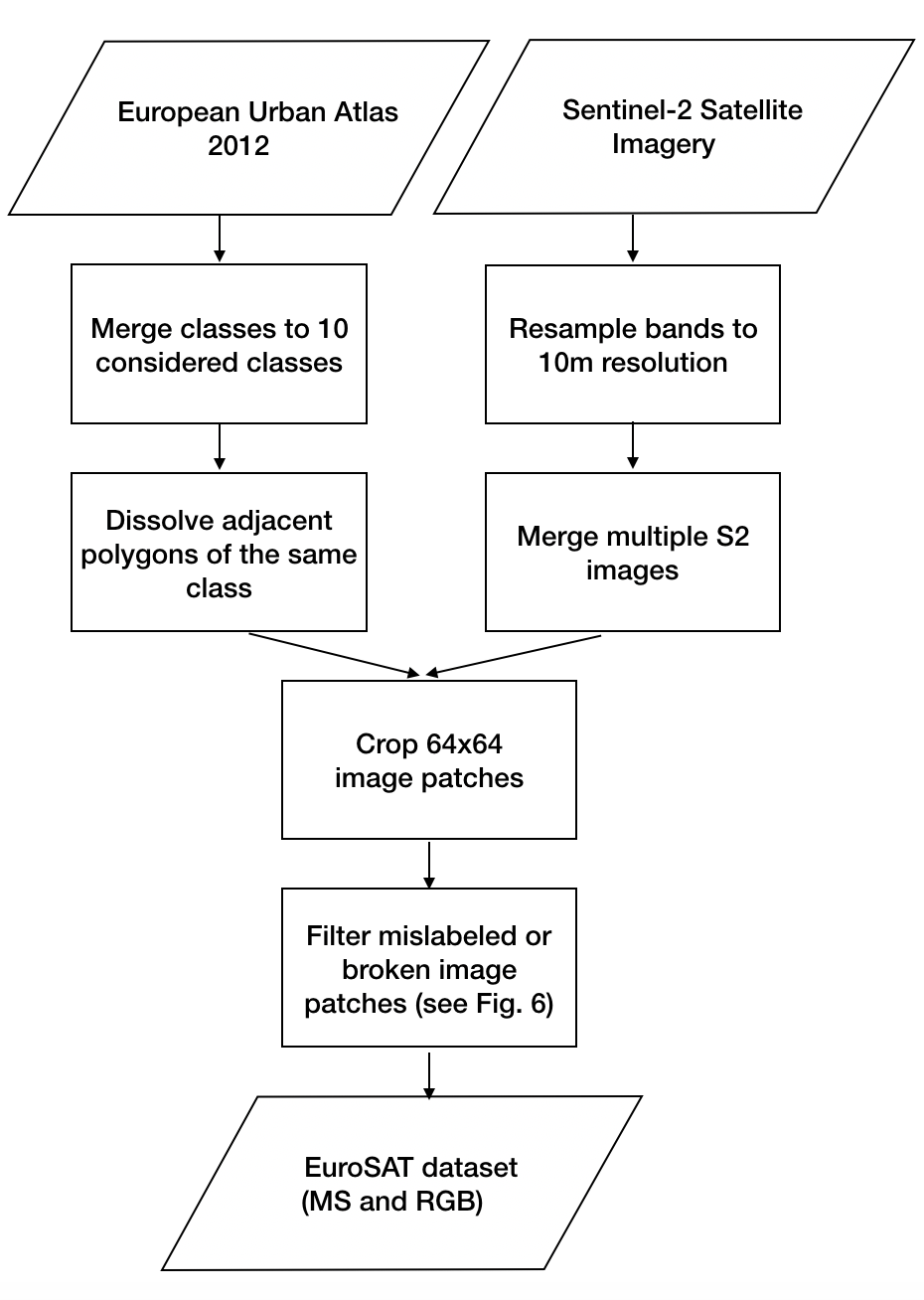}
	\caption{The diagram illustrates the EuroSAT dataset creation process.}
	\label{fig:flow_chart}
\end{figure}

\subsection{Land Use and Land Cover Classification}

Convolutional Neural Networks (CNNs) are a type of Neural Networks~\cite{lecun1989backpropagation}, which became with the impressive results on image classification challenges ~\cite{krizhevsky2012imagenet, ILSVRC15, simonyan2014very} the state-of-the-art image classification method in computer vision and machine learning.

To classify remotely sensed images, various different feature extraction and classification methods (e.g., Random Forests) were evaluated on the introduced datasets. Yang et al. evaluated Bag-of-Visual-Words (BoVW) and spatial extension approaches on the UCM dataset ~\cite{yang2010bag}. Basu et al. analyzed deep belief networks, basic CNNs and stacked denoising autoencoders on the SAT-6 dataset~\cite{basu2015deepsat}. Basu et al. also presented an own framework for the land cover classes introduced in the SAT-6 dataset. The framework extracts features from the input images, normalizes the extracted features and used the normalized features as input to a deep belief network. Besides low-level color descriptors, Penatti et al. also evaluated deep CNNs on the UCM and BCS dataset~\cite{penatti2015deep}. In addition to deep CNNs, Castelluccio et al. intensively evaluated various machine learning methods (e.g., Bag-of-Visual-Words, spatial pyramid match kernels) for the classification of the UCM and BCS dataset. 

In the context of deep learning, the used deep CNNs have been trained from scratch or fine-tuned by using a pretrained network~\cite{castelluccio2015land, nogueira2017towards, ahmad2017cnn, cheng2017remote, ma2016satellite}. The networks were mainly pretrained on the ILSVRC-2012 image classification challenge~\cite{ILSVRC15} dataset. Even though these pretrained networks were trained on images from a totally different domain, the features generalized well. Therefore, the pretrained networks proved to be suitable for the classification of remotely sensed images ~\cite{marmanis2016deep}. The presented works extensively evaluated all proposed machine learning methods and concluded that that deep CNNs outperform non-deep learning approaches on the considered datasets~\cite{castelluccio2015land, marmanis2016deep, luus2015multiview, xia2016aid}.


\begin{figure*}
\centering
  \begin{subfigure}[b]{.19\linewidth}
    \centering
    \includegraphics[width=70px]{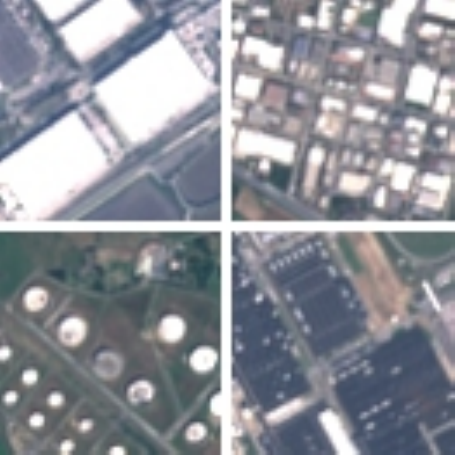}
    \caption{Industrial Buildings}\label{fig:1a}
  \end{subfigure}%
  \begin{subfigure}[b]{.19\linewidth}
    \centering
    \includegraphics[width=70px]{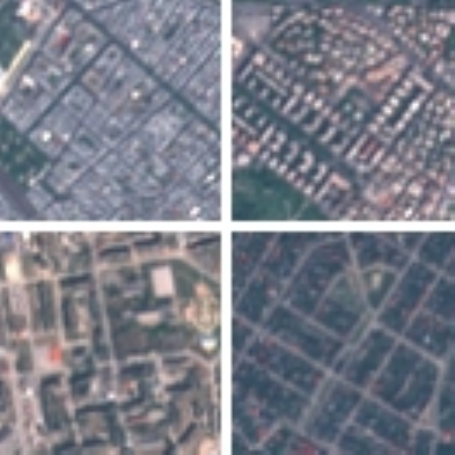}
    \caption{Residential Buildings}\label{fig:1b}
  \end{subfigure}%
  \begin{subfigure}[b]{.19\linewidth}
    \centering
    \includegraphics[width=70px]{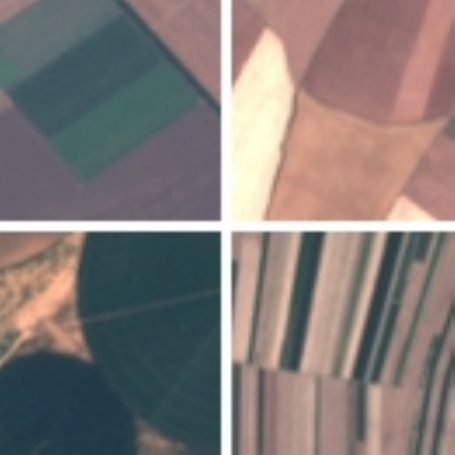}
    \caption{Annual Crop}\label{fig:1c}
  \end{subfigure}%
  \begin{subfigure}[b]{.19\linewidth}
    \centering
    \includegraphics[width=70px]{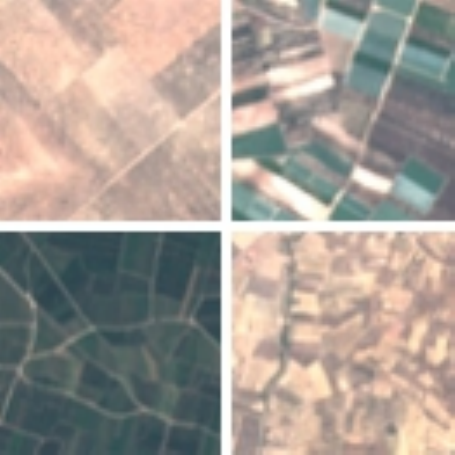}
    \caption{Permanent Crop}\label{fig:1d}
  \end{subfigure}%
  \begin{subfigure}[b]{.19\linewidth}
    \centering
    \includegraphics[width=70px]{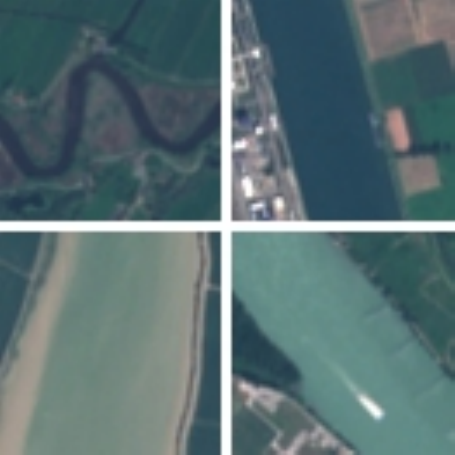}
    \caption{River}\label{fig:1i}
  \end{subfigure}%
    \vspace{+0.25cm}
  \begin{subfigure}[b]{.19\linewidth}
    \centering
    \includegraphics[width=70px]{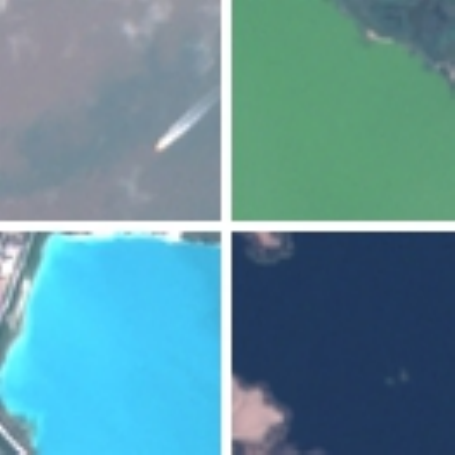}
    \caption{Sea \& Lake}\label{fig:1j}
  \end{subfigure}%
  \begin{subfigure}[b]{.19\linewidth}
    \centering
    \includegraphics[width=70px]{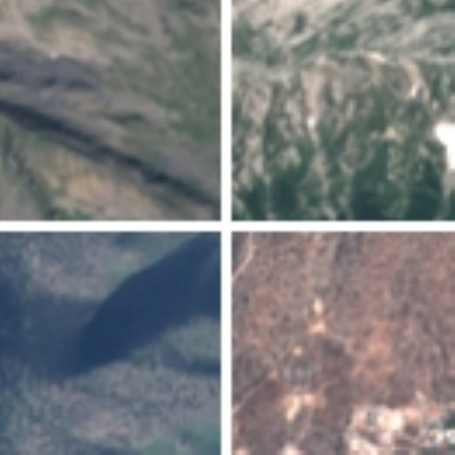}
    \caption{Herbaceous Vegetation}\label{fig:1k}
  \end{subfigure}%
  \begin{subfigure}[b]{.19\linewidth}
    \centering
    \includegraphics[width=70px]{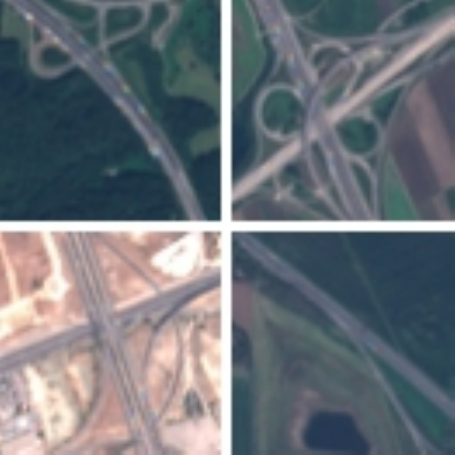}
    \caption{Highway}\label{fig:1l}
  \end{subfigure}%
  \begin{subfigure}[b]{.19\linewidth}
    \centering
    \includegraphics[width=70px]{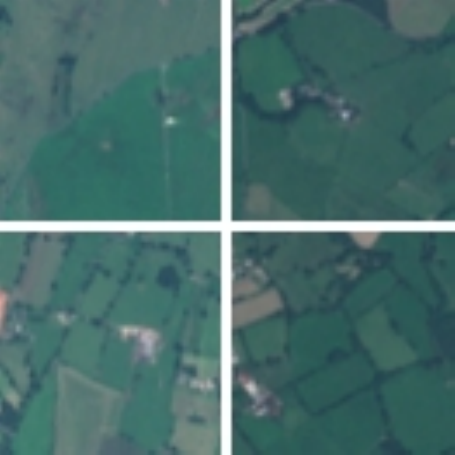}
    \caption{Pasture}\label{fig:1m}
  \end{subfigure}%
  \begin{subfigure}[b]{.19\linewidth}
    \centering
    \includegraphics[width=70px]{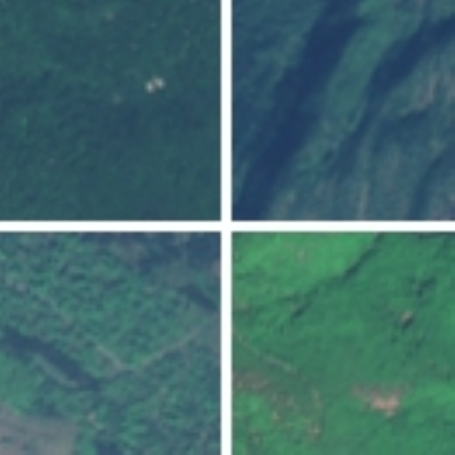}
    \caption{Forest}\label{fig:1m}
  \end{subfigure}%
  \caption{This overview shows sample image patches of all 10 classes covered in the proposed EuroSAT dataset. The images measure 64x64 pixels. Each class contains 2,000 to 3,000 image. In total, the dataset has 27,000 geo-referenced images.}\label{fig:sentinel_2dataset_labels}
\end{figure*}

\section{Dataset Acquisition}

Besides NASA with its Landsat Mission, the European Space Agency (ESA) steps up efforts to improve Earth observation within its Copernicus program. Under this program, ESA operates a series of satellites known as Sentinels. 

In this paper, we use mutli-spectral image data provided by the Sentinel-2A satellite in order to address the challenge of land use and land cover classification. Sentinel-2A is one satellite in the two-satellite constellation of the identical land monitoring satellites Sentinel-2A and Sentinel-2B. The satellites were successfully launched in June 2015 (Sentinel-2A) and March 2017 (Sentinel-2B). Both sun-synchronous satellites capture the global Earth's land surface with a Multispectral Imager (MSI) covering the 13 different spectral bands listed in Table~\ref{table:1}. The three bands B01, B09 and B10 are intended to be used for the correction of atmospheric effects (e.g., aerosols, cirrus or water vapor). The remaining bands are primarily intended to identify and monitor land use and land cover classes. In addition to mainland, large islands as well as inland and coastal waters are covered by these two satellites. Each satellite will deliver imagery for at least 7 years with a spatial resolution of up to 10 meters per pixel. Both satellites carry fuel for up to 12 years of operation which allows for an extension of the operation. The two-satellite constellation generates a coverage of almost the entire Earth's land surface about every five days, i.e. the satellites capture each point in the covered area about every five days. This short repeat cycle as well as the future availability
of the Sentinel satellites allows a continuous monitoring of the Earth's land surface for about the next 20 - 30 years. Most importantly, the data is openly and freely accessible and can be used for any application (commercial or non-commercial use). 





\begin{figure}[t!]
	\centering
	\includegraphics[width=1.0\linewidth]{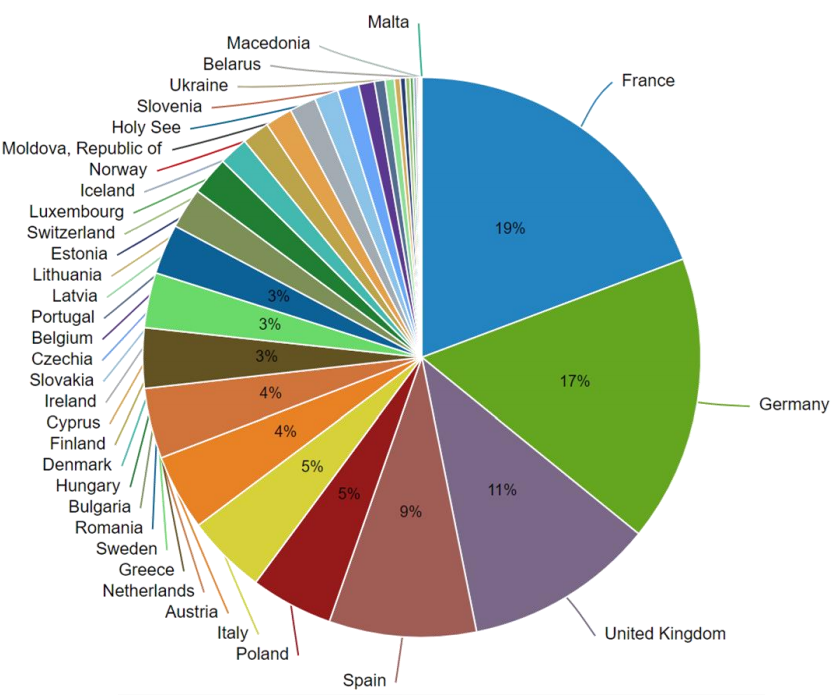}
	\caption{EuroSAT dataset distribution. The georeferenced images are distributed all over Europe. The distribution is influenced by the number of represented cities per country in the European Urban Atlas.}
	\label{fig:distribution}
\end{figure}

We are convinced that the large volume of satellite data in combination with powerful machine learning methods will influence future research. Therefore, one of our key research aims is to make this large amount of data accessible for machine learning based applications. To construct an image classification dataset, we performed the following two steps:


\begin{enumerate}
    \item Satellite Image Acquisition: We gathered satellite images of European cities distributed in over 34 countries as shown in Fig.~\ref{fig:distribution}.
    \item Dataset Creation: Based on the obtained satellite images, we created a dataset of 27,000 georeferenced and labeled image patches. The image patches measure 64x64 pixels and have been manually checked.
\end{enumerate}

\subsection{Satellite Image Acquisition}
  
We have downloaded satellite images taken by the satellite Sentinel-2A via Amazon S3. We chose satellite images associated with the cities covered in the European Urban Atlas. The covered cities are distributed over the 34 European countries: Austria, Belarus, Belgium, Bulgaria, Cyprus, Czech Republic (Czechia), Denmark, Estonia, Finland, France, Germany, Greece, Hungary, Iceland, Ireland, Italy / Holy See, Latvia, Lithuania, Luxembourg, Macedonia, Malta, Republic of Moldova, Netherlands, Norway, Poland, Portugal, Romania, Slovakia, Slovenia, Spain, Sweden, Switzerland, Ukraine and United Kingdom.

In order to improve the chance of getting valuable image patches, we selected satellite images with a low cloud level. Besides the possibility to generate a cloud mask, ESA provides a cloud level value for each satellite image allowing to quickly select images with a low percentage of clouds covering the land scene.


We aimed for the objective to cover as many countries as possible in the EuroSAT dataset in order to cover the high intra-class variance inherent to remotely sensed images. Furthermore, we have extracted images recorded all over the year to get a variance as high as possible inherent in the covered land use and land cover classes. Within one class of the EuroSAT dataset, different land types of this class are represented such as different types of forests in the forest class or different types of industrial buildings in the industrial building class. Between the classes, there is a low positive correlation. The classes most common to each other are the two presented agricultural classes and the two classes representing residential and industrial buildings. The composition of the individual classes and their relationships are specified in the mapping guide of the European Urban Atlas~\cite{mapEUA}. An overview diagram of the dataset creation process is shown in Fig.~\ref{fig:flow_chart}

\begin{table}[t!]
\vspace{+0.25cm}
\caption{All 13 bands covered by Sentinel-2's Multispectral Imager (MSI). The identification, the spatial resolution and the central wavelength is listed for each spectral band.}
\centering
\begin{tabular}{lccc}\toprule
\textbf{Band}	&\textbf{Spatial} &\textbf{Central}	\\
& \textbf{Resolution} & \textbf{Wavelength} \\
& \(m\)	& \(nm\)  \\\midrule

B01	- Aerosols	& 60	& 443	 \\
B02 - Blue 	& 10	& 490		\\
B03	- Green	& 10	& 560		\\
B04	- Red	& 10	& 665		\\
B05 - Red edge 1		& 20	& 705	 \\
B06 - Red edge 2	& 20	& 740	\\
B07	- Red edge 3	& 20	& 783	\\
B08 - NIR		& 10	& 842	\\
B08A - Red edge 4	& 20	& 865  \\
B09	- Water vapor	& 60	& 945 \\
B10	- Cirrus	& 60	& 1375 \\
B11	- SWIR 1	& 20	& 1610 \\
B12	- SWIR 2	& 20	& 2190 \\\bottomrule
\end{tabular}
\label{table:1}
\end{table}

\begin{table*}[ht!]
\centering
\caption{Classification accuracy (\%) of different training-test splits on the EuroSAT dataset.}
\begin{tabular}{|c|c|c|c|c|c|c|c|c|c|} 
    \hline
    \textbf{Method} & \textbf{10/90} & \textbf{20/80} & \textbf{30/70} & \textbf{40/60} & \textbf{50/50} & \textbf{60/40}& \textbf{70/30}& \textbf{80/20} & \textbf{90/10}\\
    \hline
    BoVW (SVM, SIFT, k = 10)& 54.54 & 56.13 & 56.77 & 57.06 & 57.22 & 57.47 & 57.71 & 58.55 & 58.44\\
    \hline
    BoVW (SVM, SIFT, k = 100)& 63.07 & 64.80 & 65.50 & 66.16 & 66.25 & 66.34 & 66.50 & 67.22 & 66.18\\
    \hline
    BoVW (SVM, SIFT, k = 500)& 65.62 & 67.26 & 68.01 & 68.52 & 68.61 & 68.74 & 69.07 & 70.05 & 69.54\\
    \hline
    CNN (two layers)& 75.88 & 79.84 & 81.29 & 83.04 & 84.48 & 85.77 & 87.24 & 87.96 & 88.66 \\
    \hline
    ResNet-50 & 75.06 & 88.53 & 93.75 & 94.01 & 94.45 & 95.26 & 95.32 & 96.43  & 96.37 \\ 
    \hline
    GoogleNet & 77.37 & 90.97 & 90.57 & 91.62 & 94.96 & 95.54 & 95.70 & 96.02 & 96.17 \\ 

    
\hline
\end{tabular}
\label{table:dataset_accuracies_splits}
\end{table*}

\subsection{Dataset Creation}

The Sentinel-2 satellite constellation provides about 1.6 TB of compressed images per day. Unfortunately, supervised machine learning is restricted even with this amount of data by the lack of labeled ground truth data. The generation of the benchmarking EuroSAT dataset was motivated by the objective of making this open and free satellite data accessible to various Earth observation applications and the observation that existing benchmark datasets are not suitable for the intended applications with Sentinel-2 satellite images.
The dataset consists of 10 different classes with 2,000 to 3,000 images per class. In total, the dataset has 27,000 images. The patches measure 64x64 pixels. We have chosen 10 different land use and land cover classes based on the principle that they showed to be visible at the resolution of 10 meters per pixel and are frequently enough covered by the European Urban Atlas to generate thousands of image patches. To differentiate between different agricultural land uses, the proposed dataset covers the classes annual crop, permanent crop (e.g., fruit orchards, vineyards or olive groves) and pastures. The dataset also discriminates built-up areas. It therefore covers the classes highway, residential buildings and industrial buildings. The residential class is created using the urban fabric classes described in the European Urban Atlas. Different water bodies appear in the classes river and sea \& lake. Furthermore, undeveloped environments such as forest and herbaceous vegetation are included. An overview of the covered classes with four samples per class is shown in Fig.~\ref{fig:sentinel_2dataset_labels}. 

\begin{figure}[t]
	\centering
	\includegraphics[width=170px]{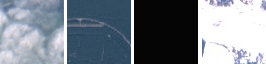}
	\caption{Four examples of bad image patches, which are intended to show industrial buildings. Clearly, no industrial building is shown due to clouds, mislabeling, dead pixels or ice/snow.}
	\label{fig:bad_patches}
\end{figure}

\begin{figure}[t]
	\centering
	\includegraphics[width=120px]{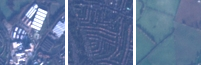}
	\caption{Color cast due to atmospheric effects.}
	\label{fig:patches_color_cast}
\end{figure}

We manually checked all 27,000 images multiple times and corrected the ground truth by sorting out mislabeled images as well as images full of snow or ice. Example images, which have been discarded, are shown in Fig.~\ref{fig:bad_patches}. The samples are intended to show industrial buildings. Clearly, no industrial building is visible. Please note, the proposed dataset has not received atmospheric correction. This can result in images with a color cast. Extreme cases are visualized in Fig.~\ref{fig:patches_color_cast}. With the intention to advocate the classifier to also learn these cases, we did not filter the respective samples and let them flow into the dataset.

Besides the 13 covered spectral bands, the new dataset has three further central innovations. Firstly, the dataset is not based on non-free satellite images like Google Earth imagery or relies on data sources which are not updated on a high-frequent basis (e.g., NAIP used in~\cite{basu2015deepsat}). Instead, an open and free Earth observation program whose satellites deliver images for the next 20 - 30 years is used allowing real-world Earth observation applications. Secondly, the dataset uses a 10 times lower spatial resolution than the benchmark dataset closest to our research but at once distinguishes 10 classes instead of 6. For instance, we split up the built-up class into a residential and an industrial class or distinguish between different agricultural land uses. Thirdly, we release the EuroSAT dataset in a georeferenced version.

With the release of the geo-referenced EuroSAT we aim to make the large amount of Sentinel-2 satellite imagery accessible for machine learning approaches. There effectiveness was successfully demonstrated in ~\cite{chen2018training, roy2018semantic, helber2018introducing}.



\section{Dataset Benchmarking}
As shown in previous work~\cite{castelluccio2015land, luus2015multiview, marmanis2016deep, nogueira2017towards}, deep CNNs have demonstrated to outperform non-deep learning approaches in land use and land cover image classification. Accordingly, we use the state-of-the-art deep CNN models GoogleNet~\cite{szegedy2015going} and ResNet-50~\cite{he2016deep, he2016identity} for the classification of the introduced land use and land cover classes. The networks make use of the inception module~\cite{szegedy2015going,szegedy2016rethinking, szegedy2016inception, lin2013network} and the residual unit~\cite{he2016deep, he2016identity}. For the proposed EuroSAT dataset, we also evaluated the performance of the 13 spectral bands with respect to the classification task. In this context, we evaluate the classification performance using single-band and band combination images.


\begin{table}[!h]
\centering
\caption{Benchmarked classification accuracy (\%) of the two best performing classifiers GoogLeNet and ResNet-50 with a 80/20 training-test split. Both CNNs have been pretrained on the image classification dataset ILSVRC-2012 ~\cite{ILSVRC15}.}
\begin{tabular}{|c|c|c|c|c|c|c|} 
    \hline
    \textbf{Method} & \textbf{UCM} & \textbf{AID} & \textbf{SAT-6} & \textbf{BCS} & \textbf{EuroSAT}\\
    \hline
    ResNet-50 & 96.42 & 94.38 & 99.56 & 93.57 & 98.57\\
    \hline
    GoogLeNet & 97.32 & 93.99 & 98.29 & 92.70 & 98.18\\

    
\hline
\end{tabular}
\label{table:dataset_accuracies}
\end{table}

\subsection{Comparative Evaluation}

We respectively split each dataset in a training and a test set (80/20 ratio). We ensured that the split is applied class-wise. While the red, green and blue bands are covered by almost all aerial and satellite image datasets, the proposed EuroSAT dataset consists of 13 spectral bands. For the comparative evaluation, we computed images in the RGB color space combining the bands red (B04), green (B03) and blue (B02). For benchmarking, we evaluated the performance of the Bag-of-Visual-Words (BoVW) approach using SIFT features and a trained SVM. In addition, we trained a shallow Convolutional Neural Network (CNN), a ResNet-50 and a GoogleNet model on the training set. We calculated the overall classification accuracy to evaluate the performance of the different models on the considered datasets. In Table~\ref{table:dataset_accuracies_splits} we show how the approaches perform in case of different training-test splits for the EuroSAT RGB dataset.

It can be seen that all CNN approaches outperform the BoVW method and, overall, deep CNNs perform better than shallow CNNs. Nevertheless, the shallow CNN classifies the EuroSAT classes with a classification accuracy of up to 89.03\%. Please note~\cite{castelluccio2015land, nogueira2017towards, sheng2012high} for the benchmarking performance of the other datasets on different training-test splits.

Table~\ref{table:dataset_accuracies} lists the achieved classification results for the two best performing CNN models GoogLeNet and ResNet-50. In this experiment, the GoogleNet and ResNet-50 CNN models were pretrained on the ILSVRC-2012 image classification dataset~\cite{ILSVRC15}. In all fine-tuning experiments, we first trained the last layer with a learning rate of 0.01. Afterwards, we fine-tuned through the entire network with a low learning rate between 0,001 and 0,0001. With a finetuned network we achieve a classification accuracy of about 2\% higher compared to randomly initialized versions of the networks which have been trained on the EuroSAT dataset with the same training-test split (see Table~\ref{table:dataset_accuracies_splits}).

\begin{figure}[t!]
	\centering
	\includegraphics[width=0.8\linewidth]{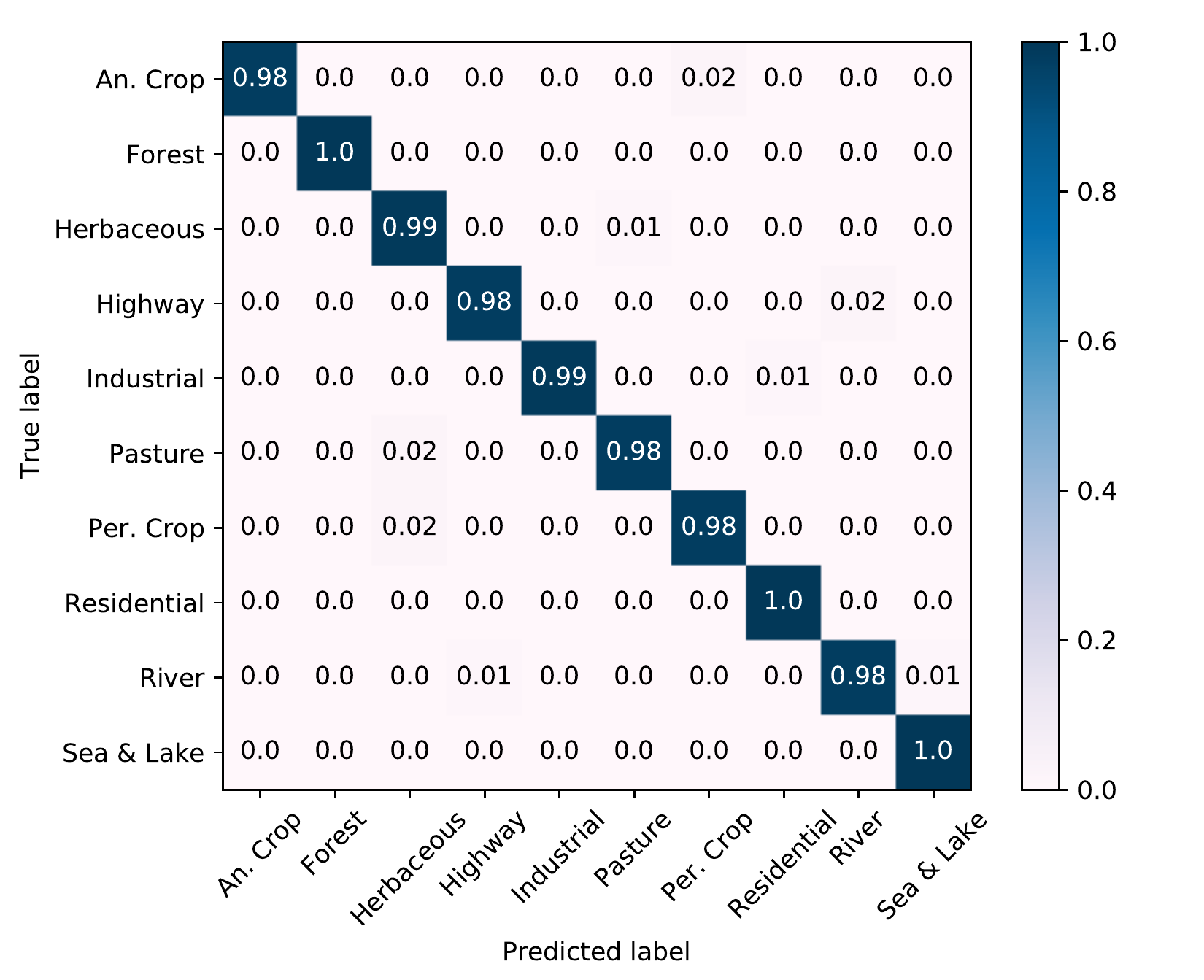}
	\caption{Confusion matrix of a fine-tuned ResNet-50 CNN on the proposed EuroSAT dataset using satellite images in the RGB color space.}
	\label{fig:confusion_matrix}
\end{figure}

The deep CNNs achieve state-of-the-art results on the UCM dataset and outperform previous results on the other three presented datasets by about 2-4\% (AID, SAT-6, BCS)~\cite{castelluccio2015land, nogueira2017towards, sheng2012high}. Table~\ref{table:dataset_accuracies} shows that the ResNet-50 architecture performs best on the introduced EuroSAT land use and land cover classes. In order to allow an evaluation on the class level, Fig.~\ref{fig:confusion_matrix} shows the confusion matrix of this best performing network. It is shown that the classifier sometimes confuses the agricultural land classes as well as the classes highway and river.



\subsection{Band Evaluation}

\begin{figure}[t!]
	\centering
	\includegraphics[width=0.7\linewidth]{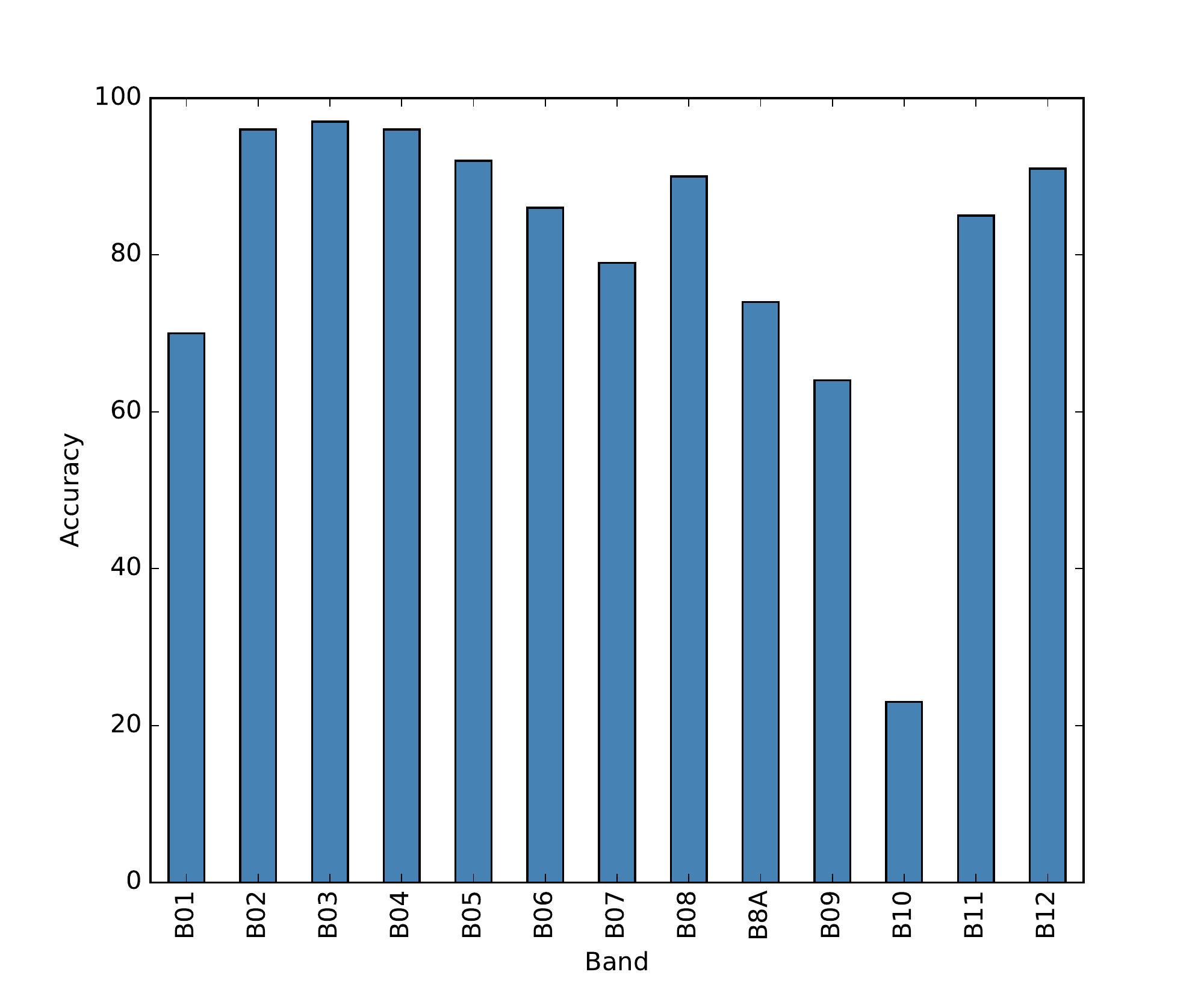}
	\caption{Overall classification accuracy (\%) of a fine-tuned ResNet-50 CNN on the given EuroSAT dataset using single-band images.}
	\label{fig:band_accuracies}
\end{figure}

In order to evaluate the performance of deep CNNs using single-band images as well shortwave-infrared and color-infrared band combinations, we used the pretrained ResNet-50 with a fixed training-test split to compare the performance of the different spectral bands. For the single-band image evaluation, we used images as input consisting of the information gathered from a single spectral band on all three input channels. We analyzed all spectral bands, even the bands not intended for land monitoring. Bands with a lower spatial resolution have been upsampled to 10 meters per pixel using cubic-spline interpolation~\cite{de1978practical}. Fig.~\ref{fig:band_accuracies} shows a comparison of the spectral band's performance. It is shown that the red, green and blue bands outperform all other bands. Interestingly, the bands red edge 1 (B05) and shortwave-infrared 2 (B12) with an original spatial resolution of merely 20 meters per pixel showed an impressive performance. The two bands even outperform the near-infrared band (B08) which has a spatial resolution of 10 meters per pixel.

\begin{table}[t!]
\centering
\vspace{+0.25cm}
\caption{Classification accuracy (\%) of a fine-tuned ResNet-50 CNN on the proposed EuroSAT dataset with the three different band combinations color-infrared (CI), shortwave-infrared (SWIR) and RGB as input.}

\begin{tabular}{|c|c|}
    \hline
    \textbf{Band Combination} & \textbf{Accuracy (ResNet-50)}\\
    \hline
    CI & 98.30\\
    \hline
    \textbf{RGB} & \textbf{98.57}\\ 
    \hline
    SWIR & 97.05\\
    \hline
\end{tabular}
\label{table:combinations_accuracies}
\end{table}

In addition to the RGB band combination, we also analyzed the performance of the shortwave-infrared and color-infrared band combination. Table~\ref{table:combinations_accuracies} shows a comparison of the performance of these combinations. As shown, band combination images outperform single-band images. Furthermore, images in the RGB color space performed best on the introduced land use and land cover classes. Please note, networks pretrained on the ILSVRC-2012 image classification dataset have initially not been trained on images other than RGB images.

\section{Applications}
The openly and freely accessible satellite images allow a broad range of possible applications. In this section, we demonstrate that the novel dataset published with this paper allows real-world applications. The classification result with an overall accuracy of 98.57\% paves the way for these applications. We show land use and land cover change detection applications as well as how the the trained classifier can assist in keeping geographical maps up-to-date.

\subsection{Land Use and Land Cover Change Detection}

\begin{figure}[h!]
	\centering
	\includegraphics[width=0.8\linewidth]{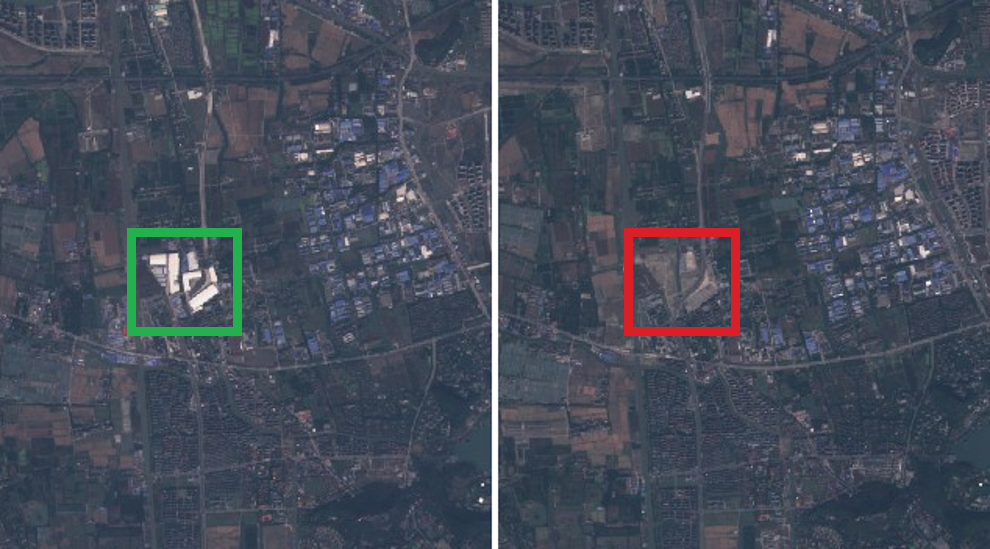}
	\caption{The left image was acquired in the surroundings of Shanghai in December 2015 showing an area classified as industrial. The right image shows the same region in December 2016 revealing that the industrial buildings have been demolished.}
	
	\label{fig:changeDetection_industrial}
\end{figure}

\begin{figure}
	\centering
	\includegraphics[width=0.8\linewidth]{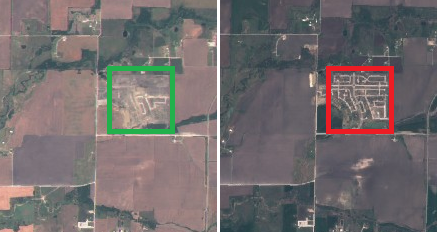}
	\caption{The left image was acquired in the surroundings of Dallas, USA in August 2015 showing no dominant residential buildings in the highlighted area. The right image shows the same area in March 2017 showing that residential buildings have been built up.}
	
	\label{fig:changeDetection_residential}
\end{figure}

\begin{figure}[h!]
	\centering
	\includegraphics[width=0.8\linewidth]{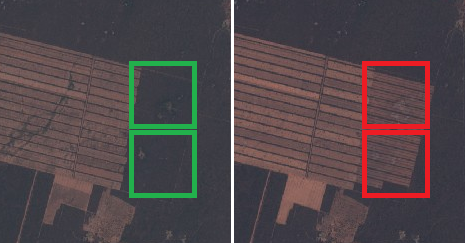}
	\caption{The left image was acquired near Villamontes, Bolivia in October 2015. The right image shows the same area in September 2016 revealing that a large land area has been deforested.}
	\label{fig:changeDetection_deforestation}
\end{figure}

Since the Sentinel-2 satellite constellation will scan the Earth's land surface for about the next 20 - 30 years on a repeat cycle of about five days, a trained classifier can be used for monitoring land surfaces and detect changes in land use or land cover. To demonstrate land use and land cover change detection, we selected images from the same spatial region but from different points in time. Using the trained classifier, we analyzed 64x64 image regions. A change has taken place if the classifier delivers different classification results for patches taken from the same spatial 64x64 region. In the following, we show three examples of spotted changes. In the first example shown in Fig.~\ref{fig:changeDetection_industrial}, the classification system recognized that the land cover has changed in the highlighted area. The left image was acquired in the surroundings of Shanghai, China in December 2015 showing an area classified as industrial. The right image shows the same area in December 2016 revealing that the industrial buildings have been demolished. The second example is illustrated in Fig.~\ref{fig:changeDetection_residential}. The left image was acquired in the surroundings of Dallas, USA in August 2015 showing no dominant residential buildings in the highlighted area. The right image shows the same area in March 2017. The system has identified a change in the highlighted area revealing that residential buildings have been constructed. The third example presented in Fig.~\ref{fig:changeDetection_deforestation} shows that the system detected deforestation near Villamontes, Bolivia. The left image was acquired in October 2015. The right image shows the same region in September 2016 revealing that a large area has been deforested. The presented examples are particularly of interest in urban area development, nature protection or sustainable development. For instance, deforestation is a main contributor to climate change, therefore the detection of deforested land is of particular interest (e.g., to notice illegal clearing of forests).


\subsection{Assistance in Mapping}
While a classification system trained with 64x64 image patches does not allow a finely graduated per-pixel segmentation, it cannot only detect changes as shown in the previous examples, it can also facilitate keeping maps up-to-date. This is an extremely helpful assistance with maps created in a crowdsourced manner like OpenStreetMap (OSM). A possible system can verify already tagged areas, identify mistagged areas or bring large area tagging. The proposed system is based on the trained CNN classifier providing a classification result for each image patch created in a sliding windows based manner.

\begin{figure*}
	\centering
	\includegraphics[width=0.9\linewidth]{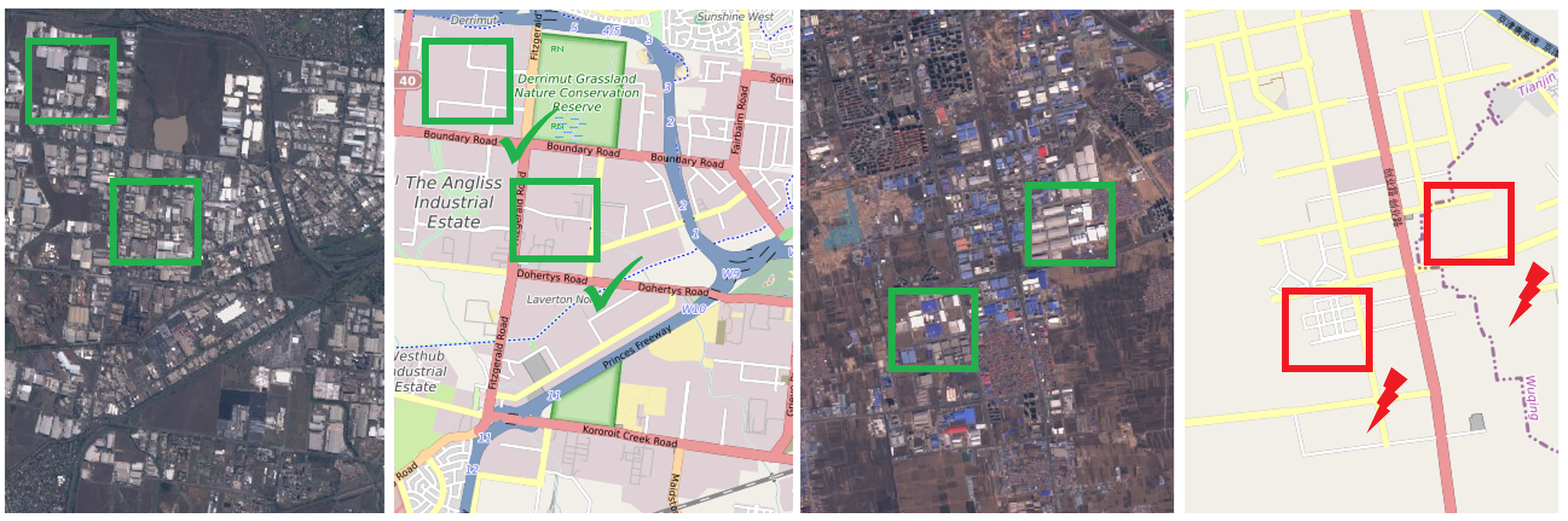}
	\caption{A patch-based classification system can verify already tagged areas, identify mistagged areas or bring large area tagging as shown in the above images and maps. The left Sentinel-2 satellite image was acquired in Australia in March 2017. The right satellite image was acquired in the surroundings of Shanghai, China in March 2017. The corresponding up-to-date OpenStreetMap (OSM) mapping images show that the industrial areas in the left satellite image are almost completely covered (colored gray). However, the industrial areas in the right satellite image are not properly covered.}
\label{fig:changeDetection_osm}
\end{figure*}

As shown in Fig.~\ref{fig:changeDetection_osm}, the industrial buildings seen in the left up-to-date satellite image are almost completely covered in the corresponding OSM mapping. The right up-to-date satellite image also shows industrial buildings. However, a major part of the industrial buildings is not covered in the corresponding map. Due to the high temporal availability of Sentinel-2 satellite images in the future, this work together with the published dataset can be used to build systems which assist in keeping maps up-to-date. A detailed analysis of the respective land area can then be provided using high-resolution satellite images and an advanced segmentation approach~\cite{bischke2017multitask_buildings, Kampffmeyer_2016_CVPR_Workshops}.

\section{Conclusion}
In this paper, we have addressed the challenge of land use and land cover classification. For this task, we presented a novel dataset based on remotely sensed satellite images. To obtain this dataset, we have used the openly and freely accessible Sentinel-2 satellite images provided in the Earth observation program Copernicus. The proposed dataset consists of 10 classes covering 13 different spectral bands with in total 27,000 labeled and geo-referenced images. We provided benchmarks for this dataset with its spectral bands using state-of-the-art deep Convolutional Neural Network (CNNs). For this novel dataset, we analyzed the performance of the 13 different spectral bands. As a result of this evaluation, the RGB band combination with an overall classification accuracy of 98.57\% outperformed the shortwave-infrared and the color-infrared band combination and leads to a better classification accuracy than all single-band evaluations. Overall, the available free Sentinel-2 satellite images offer a broad range of possible applications. This work is a first important step to make use of the large amount of available satellite data in machine learning allowing to monitor Earth's land surfaces on a large scale. The proposed dataset can be leveraged for multiple real-world Earth observation applications. Possible applications are land use and land cover change detection or the improvement of geographical maps.

\appendices
\section*{Acknowledgment}
This work was partially funded by the BMBF project DeFuseNN (01IW17002). The authors thank NVIDIA for the support within the NVIDIA AI Lab program.

\ifCLASSOPTIONcaptionsoff
  \newpage
\fi

\end{document}